\pdfoutput=1

\documentclass[11pt]{article}

\usepackage[final]{acl}
\usepackage[most]{tcolorbox}
\usepackage{times}
\usepackage{latexsym}
\usepackage{amsmath}
\usepackage{booktabs}
\usepackage{arydshln}
\usepackage{subcaption}
\newenvironment{sizeddisplay}[1]
 {\par\nopagebreak#1\noindent\ignorespaces}
 {\nopagebreak\ignorespacesafterend}
\usepackage[T1]{fontenc}

\usepackage[utf8]{inputenc}

\usepackage{microtype}

\usepackage{inconsolata}

\usepackage{multirow}
\usepackage{multicol}
\usepackage{xspace}
\usepackage{graphicx}
\usepackage{amsmath, amsfonts, amssymb}
\newcommand{\synthiabold}{\textbf{\textsc{Synthia}}\xspace}
\newcommand{\synthia}{\textsc{Synthia}\xspace}
\NewDocumentCommand{\heng}
{ mO{} }{\textcolor{red}{\textsuperscript{\textit{Heng}}\textsf{\textbf{\small[#1]}}}}
\NewDocumentCommand{\kw}
{ mO{} }{\textcolor{blue}{\textsuperscript{\textit{KW}}\textsf{\textbf{\small[#1]}}}}
\NewDocumentCommand{\violet}
{ mO{} }{\textcolor{purple}{\textsuperscript{\textit{Violet}}\textsf{\textbf{\small[#1]}}}}
\NewDocumentCommand{\ember}
{ mO{} }{\textcolor{blue}{\textsuperscript{\textit{Ember}}\textsf{\textbf{\small[#1]}}}}
\NewDocumentCommand{\xiaomeng}
{ mO{} }{\textcolor{green}{\textsuperscript{\textit{Xiaomeng}}\textsf{\textbf{\small[#1]}}}}
\NewDocumentCommand{\jeongh}
{ mO{} }{\textcolor{purple}{\textsuperscript{\textit{Jeonghwan}}\textsf{\textbf{\small[#1]}}}}

\NewDocumentCommand{\Jiateng}
{ mO{} }{\textcolor{orange}{\textsuperscript{\textit{Jiateng}}\textsf{\textbf{\small[#1]}}}}
\NewDocumentCommand{\zhenhailong}
{ mO{} }{\textcolor{teal}{\textsuperscript{\textit{Zhenhailong}}\textsf{\textbf{\small[#1]}}}}
\title{\synthia: Novel Concept Design with Affordance Composition}

\author{%
 Hyeonjeong Ha$^{1}$\thanks{~~Equal contribution.}, Xiaomeng Jin$^{1*}$,  Jeonghwan Kim$^{1}$, \\  
 \bf Jiateng Liu$^{1}$, Zhenhailong Wang$^{1}$, Khanh Duy Nguyen$^{1}$, Ansel Blume$^{1}$, \\
 \bf  Nanyun Peng$^{2}$, Kai-Wei Chang$^{2}$, Heng Ji$^{1}$\\
$^{1}$University of Illinois Urbana-Champaign\\
$^{2}$University of California Los Angeles \\
\texttt{\{hh38, xjin17, hengji\}@illinois.edu}  \\
}

\begin{document}
\maketitle
\begin{abstract}
Text-to-image (T2I) models enable rapid concept design, making them widely used in AI-driven design. While recent studies focus on generating semantic and stylistic variations of given design concepts, \textit{functional coherence}---the integration of multiple affordances into a single coherent concept---remains largely overlooked. In this paper, we introduce 
\synthiabold, a framework for generating visually novel and functionally coherent designs based on desired affordances. Our approach leverages a hierarchical concept ontology that decomposes concepts into parts and affordances, serving as a crucial building block for functionally coherent design. We also develop a curriculum learning scheme based on our ontology that contrastively fine-tunes T2I models to progressively learn affordance composition while maintaining visual novelty. To elaborate, we (i) gradually increase affordance distance, guiding models from basic concept-affordance association to complex affordance compositions that integrate parts of distinct affordances into a single, coherent form, and (ii) enforce visual novelty by employing contrastive objectives to push learned representations away from existing concepts. Experimental results show that \synthia outperforms state-of-the-art T2I models, demonstrating absolute gains of 25.1\% and 14.7\% for novelty and functional coherence in human evaluation, respectively. Code is available at \href{https://github.com/HyeonjeongHa/SYNTHIA}{https://github.com/HyeonjeongHa/SYNTHIA}.


\end{abstract}

\section{Introduction}


Imagine a coffee machine with wheels that brews your morning coffee and delivers it directly to your bed. This example illustrates a novel concept that is atypical and dissimilar to everyday objects we regularly encounter. Such \textit{novel concept design} is defined as the synthesis of concepts that are visually unique and functionally coherent through the effective fusion of disparate concepts (e.g., \texttt{coffee machine}, \texttt{trolley}) based on the user’s desired affordances (e.g., \texttt{brew}, \texttt{deliver}). This process mirrors how humans blend ideas across cognitive domains to generate creative innovations~\cite{fauconnier2002wwt, Han2018THECD}.


\begin{figure*}[t]
\centering
\begin{subfigure}[t]{0.46\linewidth}
    \centering
    \includegraphics[width=\linewidth]{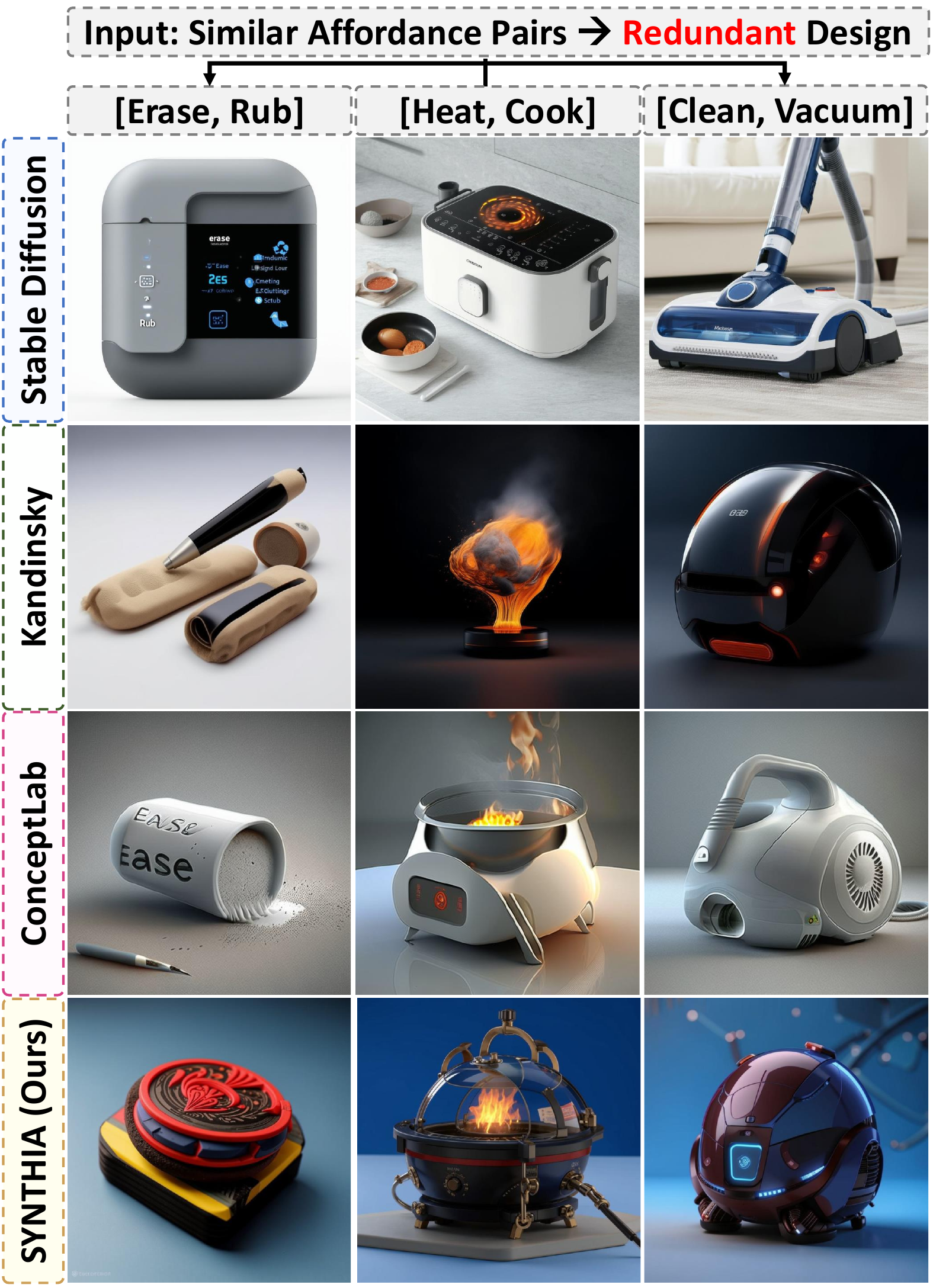}
    \caption{Generated concepts with similar affordances.}
    \label{fig:close}
\end{subfigure}
\hspace{0.1in}
\begin{subfigure}[t]{0.46\linewidth}
    \centering
    \includegraphics[width=\linewidth]{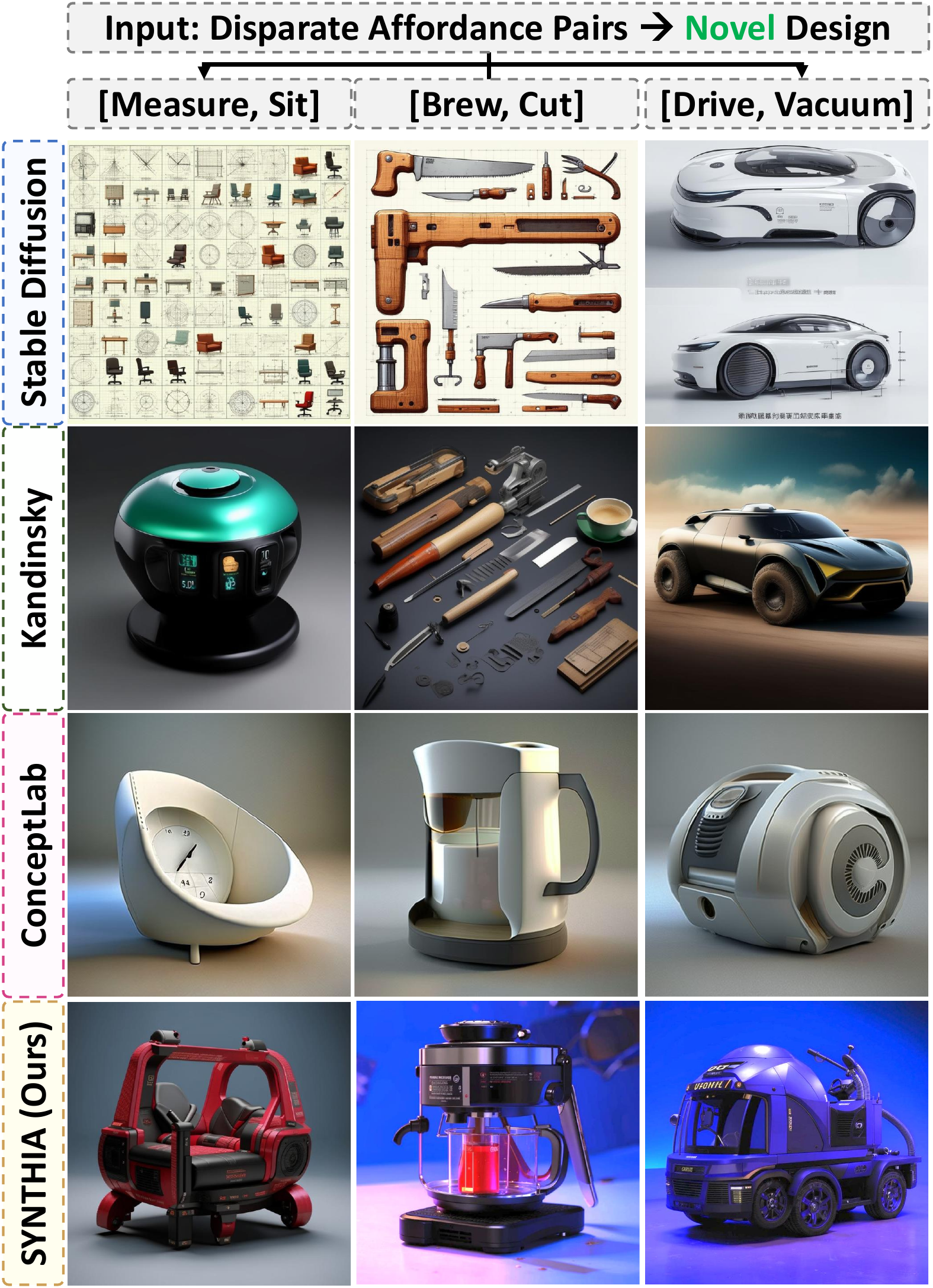}
    \caption{Generated concepts with distant affordances.}
    \label{fig:distant}
\end{subfigure}
\vspace{-0.05in}
\caption{\textbf{Effect of Affordance Sampling on Novel Concept Generation.} Our affordance sampling strategy selects disparate affordance pairs within our ontology, promoting novel functional coherence rather than redundant combinations. Baseline models tend to generate existing concepts for close affordances (Fig.~\ref{fig:close}) but struggle with distant pairs, often introducing multiple objects or omitting functions (Fig.~\ref{fig:distant}). In contrast, our models consistently generate functionally coherent novel concepts, achieving higher novelty scores for distant affordance pairs.}
\label{fig:examples}
\end{figure*}

Existing studies on conceptual design using T2I models have enabled rapid ideation of novel visual concepts \citep{cai2023designaid, ma2023conceptual, wang2024inspired, lin2025inkspire} by identifying user challenges such as interpreting abstract concepts in language to help visualize a novel design concept~\citep{lin2025inkspire}, or using large language models (LLMs) to bootstrap initial ideation in texts \citep{cai2023designaid, zhu2023generative}. However, they often naively feed LLM-generated textual prompts into T2I models, relying on simple key phrases or semantic variations of concept description~\citep{cai2023designaid, wang2024inspired}. While existing works show that T2I models can generate images that seem to correctly reflect complex human-formulated textual descriptions (e.g., \textit{``beautiful rendering of neon lights in futuristic cyberpunk city''}), they do not focus on whether a model can synthesize a novel concept when given a set of affordances as input, while ensuring these affordances are preserved.


An important aspect lacking in existing approaches to concept synthesis is their focus on pixel-based control, overlooking the structural and functional roles embedded in design. Many real-world concepts are naturally ``decomposable'' into parts, where each part signals a specific functionality. To address this, we propose \synthiabold, a framework for Concept \textbf{Synth}es\textbf{i}s with \textbf{A}ffordance composition that generates functionally coherent and visually novel concepts given a set of desired affordances. Unlike prior works relying on complex descriptive text to generate stylistic variations or aesthetic features~\cite{richardson2024conceptlab, vinker2023concept}, \synthia leverages affordances--defined as ``the functionality offered by an object or its parts''---as a structural guide for novel concept synthesis. By aligning textual descriptions with affordances as control signals, our models implicitly learn to ``decompose and reassemble'' functional parts, ensuring that, for instance, a hybrid of a \texttt{coffee machine} and a \texttt{trolley} not only appears novel but also retains its brewing and mobility functions, achieving \textit{functional coherence}.

To facilitate structured affordance composition, we construct a hierarchical concept ontology that decomposes visual concepts (e.g., \texttt{Sofa}) into their constituent parts (e.g., \texttt{leg}, \texttt{cushion}) and associated affordances (e.g., \texttt{support}, \texttt{rest}). It provides a structured representation of concept-affordance associations, serving as the foundation for generating functionally meaningful designs. Inspired by the theory of combinational creativity in humans~\cite{Han2018THECD}, which suggests novel concepts emerge from disparate ideas, we propose an affordance sampling mechanism that strategically selects affordances associated with sufficiently different concepts using our novel similarity-based metric (\S\ref{sec:data_gen}). This ensures that generated designs integrate novel functionalities, avoiding trivial combinations, whereas random sampling yields similar affordances (e.g., \texttt{cook}, \texttt{heat}) that result in redundant outputs (Fig.~\ref{fig:close}). \looseness=-1

We also introduce a new curriculum learning scheme that fine-tunes T2I models to progressively learn affordance composition while maintaining visual novelty. Our curriculum gradually increases the affordance distance, allowing models to first learn basic concept-affordance associations from close affordance pairs before tackling complex affordance compositions that integrate multiple affordances into a single, coherent form. To further ensure novelty, we employ contrastive objectives to push learned representations away from existing concepts in our ontology. This addresses a critical limitation of existing T2I models, which struggle to generate coherent multi-functional concepts (Fig.~\ref{fig:distant}). Without structured affordance composition, models tend to default to familiar objects---e.g., when prompted with \texttt{drive} and \texttt{vacuum} affordances, Stable Diffusion models simply generate a car with missing \texttt{vacuum} functions (Fig.~\ref{fig:distant}), rather than blending both affordances into an integrated design. Importantly, unlike existing AI-driven design frameworks that rely on detailed LLM-generated descriptions, \synthia enables direct affordance-based prompting, e.g., ``\textit{a new design that has functions of \{desired affordances\}.}''. Our model implicitly learns concept-affordance associations, producing novel, structured designs without redundant textual prompting. \looseness=-2

To evaluate our framework, we uniformly sample 500 unseen affordance pairs from our ontology and assess generated concepts using automatic and human evaluation. We design evaluation metrics (\S\ref{sec:eval}) that measure faithfulness, novelty, practicality, and coherence.  
Experiments show that \synthia significantly outperforms baselines, creating designs that are visually novel and functionally coherent, with consistently higher scores across all metrics. Our contributions are as follows:
\begin{itemize}
    \item We introduce a hierarchical concept ontology that encodes concept-affordance associations, serving as crucial building blocks for novel concept synthesis with functional coherence.
    \item We propose an affordance sampling strategy that guides disparate affordance selection, avoiding redundant functionalities while ensuring coherent concept synthesis.
    \item We develop a curriculum-based optimization for affordance composition that fine-tunes T2I models, enabling T2I models to fuse multiple affordances into a single coherent concept.
    
\end{itemize}

\section{Related Work}
\paragraph{\textbf{Text-to-Image Models.}}
The advancement of text-to-image (T2I) models has enabled high-quality image synthesis from textual descriptions \citep{sohn2023styledrop, xue2024raphael, shi2024instantbooth, chen2024pixart, zhou2025contrastive}. Especially, the invention of diffusion-based models, such as DALL-E \citep{dalle} and Stable Diffusion \citep{stabled}, significantly increases the performance of the T2I generation by utilizing a transformer-based architecture, where the image embeddings and text encodings are aligned in the shared representation space. For instance, \citet{bao2024separateandenhancecompositionalfinetuningtext2image} propose a compositional fine-tuning method for T2I Diffusion Models that focuses on two novel objectives and performs fine-tuning on critical parameters. However, these models still struggle to understand practical functionalities and integrate multiple components into coherent novel concepts. This highlights the need for a new framework to enhance the compositional reasoning ability of T2I models, which our work aims to address.
\paragraph{\textbf{Novel Concept Generation.}}
The great power of T2I models provides a potential boost to content creation \citep{ko2023large, rangwani2024crafting, sankar2024novelideagenerationtool, rahman2024visual, tang2024lgm, zhou2025contrastivevisualdataaugmentation}. Novel concept generation aims to produce visual outputs that extend the existing concepts by specifying the requirements as input to the T2I models. For instance, Concept Weaver \citep{kwon2024concept} first generates a template image based on a text prompt, then refines it using a concept fusion strategy. ConceptLab \citep{richardson2024conceptlab} utilizes Diffusion Prior models and formulates the generation process as an optimization process over the output space of the diffusion prior. Yet, they focus on concept-level generation and ignore the relationships between concepts and their parts. By prioritizing aesthetics, they limit real-world practicality. Our work bridges this gap by designing an affordance-driven novel concept synthesis framework, integrating desired functions to output novel but practical concepts.


\section{\synthia: Novel Concept Design with Affordance Composition}
\begin{figure*}[t]
    \centering
    \includegraphics[width=\linewidth]{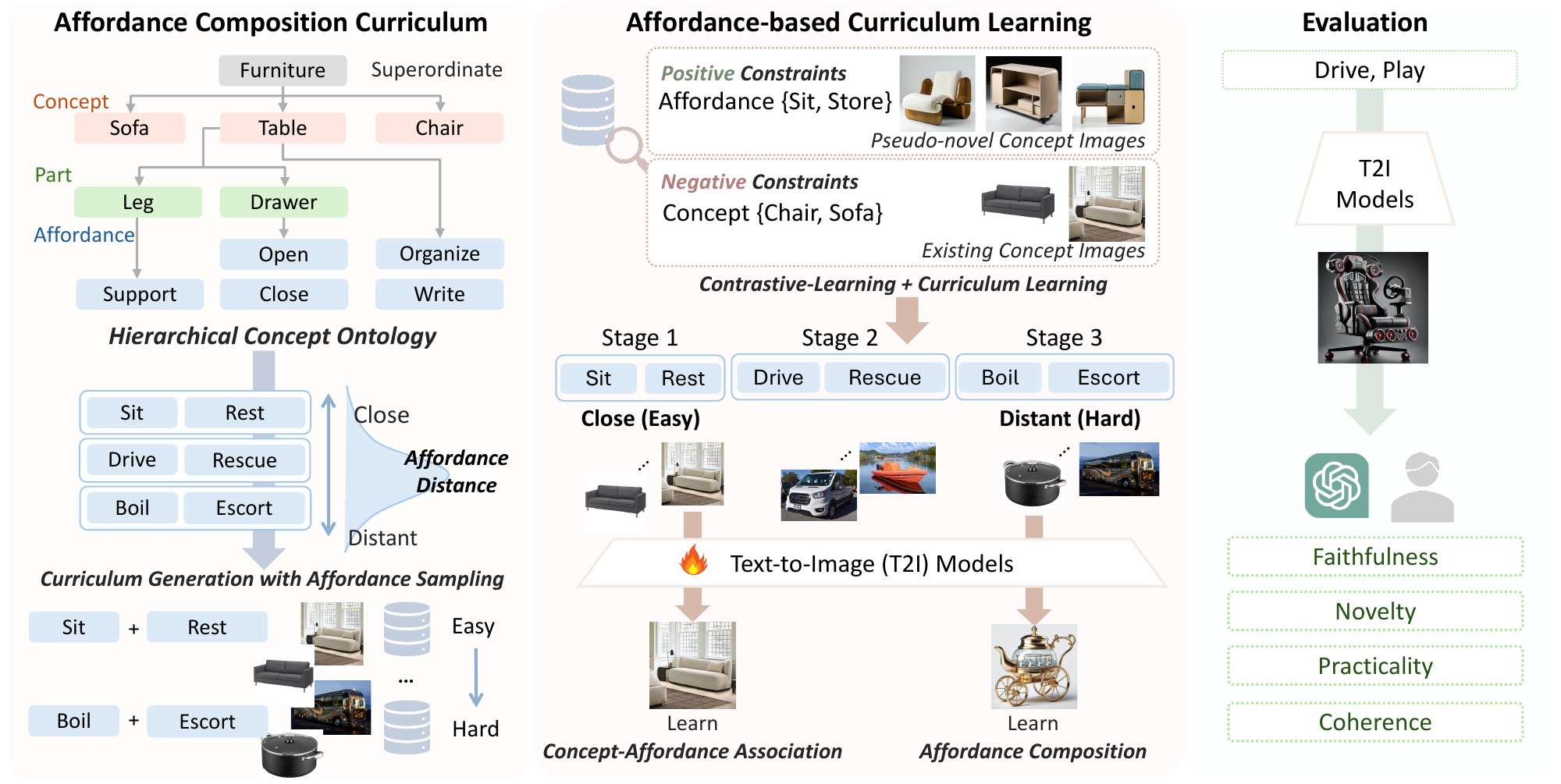}
    \caption{\textbf{\synthia: Novel Concept Design with Affordance Composition.} \synthia comprises three stages: (1) Affordance composition curriculum construction, (2) Affordance-based curriculum learning, and (3) Evaluation. In the first stage, we build a training curriculum through sampling affordance pairs from our ontology by gradually increasing the affordance distances. Using our curriculum, we fine-tune T2I models, where they first learn concept-affordance associations from easy data, then integrate multiple affordances into a single functional form from hard data. We employ a contrastive objective with positive (affordances), negative (concepts) constraints, and corresponding images, enforcing visual novelty different from existing concepts. Finally, we evaluate models through automatic evaluation and human evaluation with four metrics: faithfulness, and novelty, practicality, coherence. \looseness=-1}
    \label{fig:main_figure}
\end{figure*}

Our ultimate goal is to utilize T2I models in designing novel concepts that are both visually novel and functionally coherent. Specifically, we take the desired affordances as text inputs, the T2I models should generate an image that depicts the novel concept design.
To achieve this, we (1) construct a training recipe that explicitly embeds hierarchical relations on visual concepts, parts, and corresponding affordances, and (2) fine-tune T2I models with curriculum-based optimization (Fig.~\ref{fig:main_figure}).

\subsection{Affordance Composition Curriculum}
\label{sec:data_gen}
The primary challenge in the novel concept generation of existing T2I models is the lack of structured functional grounding. These models often struggle to design visually novel yet functionally coherent concepts while maintaining intended functionalities. For example, when combining affordances like \texttt{Brew} and \texttt{Cut}, they may prioritize aesthetics over functionality, omitting parts or objects relevant to \texttt{Brew} (Fig.~\ref{fig:distant}). To address this, we construct a structured training recipe in two key steps: (1) building a hierarchical concept ontology, and (2) designing an affordance sampling strategy for curriculum-based training. This improves the model's composition ability by learning the connection between concepts and affordances.

\paragraph{Hierarchical Concept Ontology.}
To provide a structured basis for novel concept synthesis with functional coherence, we define a hierarchical concept ontology that decomposes visual concepts into constituent parts and their affordances, capturing concept-affordance associations (Fig.~\ref{fig:ontology}). This ontology allows T2I models to retrieve relevant parts based on affordances, enabling generation to be well-grounded on the functionality of concepts rather than superficial visual feature combinations. Formally, we define the ontology as a four-level hierarchy $\mathcal{O} = (\mathcal{S}, \mathcal{C}, \mathcal{P}, \mathcal{A})$. The superordinate $\mathcal{S}$ denotes the highest-level categories, such as \texttt{furniture}, followed by Concept $\mathcal{C}$, which is the set of visual concepts. Each $c \in \mathcal{C}$ belongs to a superordinate category $s \in \mathcal{S}$, e.g., $\mathcal{S}_\texttt{table} = \texttt{furniture}$, and decomposes into its parts $\mathcal{P}$ that serve specific functions in an object design, e.g., $\mathcal{P}_{\texttt{table}} = \{\texttt{leg}, \texttt{drawer}\}$. The Affordance $\mathcal{A}$ describes functionalities of concepts and parts. Both a concept $c \in C$ and its part $p \in P$ are linked to affordances set $\mathcal{A}_c = \{a_1, \cdots, a_n\} \in \mathcal{A}$, e.g, $\mathcal{A}_{\texttt{table}} = \{\texttt{write}, \texttt{organize}\}$, and $\mathcal{A}_p = \{p_1, \cdots, p_n\}$, e.g., $\mathcal{A}_{\texttt{leg}} = \{\texttt{support}\}$. Our ontology spans 30 superordinates, 590 concepts, 1172 parts, and 686 affordances, explicitly providing a structured representation of how affordance is associated with fine-grained parts for functionally grounded novel concept synthesis.

\paragraph{Affordance Sampling.}
\label{sec:affor-sampling}
Given our ontology, we can utilize it to create fine-tuning data to improve the functional coherence of the novel concept generated by the T2I models. A naive approach to obtaining training data would be to exhaustively pair all possible affordances. 
However, this would yield 235K affordance pairs, which is computationally expensive. Moreover, random combination risks generating redundant concepts (e.g., \texttt{Heat} and \texttt{Cook} examples in Fig.~\ref{fig:close}) or functionally incoherent objects. To achieve sufficiently different affordance pairs that enable novel concept synthesis while still being functionally integrable, we introduce a distance-based affordance sampling strategy that selects meaningful, disparate affordance pairs based on ontology-derived distances.

We define a concept distance $D_{C}(c_i, c_j)$ between two concepts $c_i, c_j \in \mathcal{C}$ by incorporating functional relatedness at the affordance level and semantic similarity at the concept level. We compute functional relatedness using Jaccard similarity $J(X, Y) = |X \cap Y| / |X \cup Y|$ between their affordance sets while quantifying semantic similarity $\text{Sim}$ by measuring embedding similarity using the BERT~\citep{devlin2019bertpretrainingdeepbidirectional} model as follows:
\begin{align}
    \label{eq:dist_concept}
     D_{\mathcal{C}} (c_i, c_j) &= \alpha*\{J(\mathcal{A}_{c_i}, \mathcal{A}_{c_j}) + J(\mathcal{A}_{P_{c_i}}, \mathcal{A}_{P_{c_j}})\} \notag \\
                        &\quad+ \beta*\text{Sim}(\text{BERT}(c_i), \text{BERT}(c_j)),  
\end{align}
where $\alpha, \beta$ are adjustable hyperparameters that balance between functional relatedness based on affordances, and semantic relevance of concepts, respectively.
Since we prioritize affordance-level similarity over concept-level similarity during training, we set $\alpha=0.7$ and $\beta=0.3$.
Two semantically similar concepts sharing more affordances have closer distances, such as \texttt{sofa} and \texttt{chair}, while those that have different affordances and semantic differences, such as \texttt{car} and \texttt{vacuum cleaner} have more distant distances.


We further obtain the affordance distance $D_{\mathcal{A}} (a_i, a_j)$ between two affordances $a_i, a_j \in \mathcal{A}$ by averaging the pairwise concept distances $D_{\mathcal{C}}(\cdot, \cdot)$ between associated concepts:
\begin{sizeddisplay}{\normalsize}
\begin{align}
     \label{eq:dist_affordance}
     D_{\mathcal{A}} (a_i, a_j) &= \frac{1}{|C_{a_i}| \cdot |C_{a_j}|} \sum_{c_p \in C_{a_i}} \sum_{c_q \in C_{a_j}} D_{C}(c_p, c_q),
\end{align}
\end{sizeddisplay}
where $C_{a_i}$ and $C_{a_j}$ are the sets of concepts associated with affordances $a_i$ and $a_j$, respectively. The resulting $D_{\mathcal{A}}(\cdot, \cdot)$ is distributed from 0.1 to 1.0.

Based on our distance metric, close affordance pairs associated with similar concepts, e.g., $\{\texttt{sit}, \texttt{rest}\}$ from $\{\texttt{sofa}, \texttt{chair}\}$, support learning basic affordance-concept associations, which can be easily merged into existing concepts. In contrast, distant affordance pairs derived from sufficiently distant concepts, e.g., $\{\texttt{drive}, \texttt{vacuum}\}$ from $\{\texttt{car}, \texttt{vacuum cleaner}\}$, enforce greater functional coherence by requiring meaningful part-affordance integration, which is more complex than a trivial combination. 
\paragraph{Curriculum Construction.}
In novel concept generation, existing T2I models struggle with (1) concept-affordance associations and (2) the composition of functionally coherent affordances into a single concept. To address these challenges with limited data, we propose a three-stage curriculum that progressively increases affordance pair distances. In the earliest stage, we utilize close affordance pairs to reinforce fundamental knowledge of the concept-affordance associations. The second stage employs the affordance pairs from the mid-range distances to encourage the model to learn the fine-grained compositional structures while maintaining prior knowledge. In the last stage, we only introduce distant affordance pairs to challenge the model to synthesize novel, functionally coherent concepts by applying the previously learned basics on the fine-grained parts and affordance relations.

We sample 600 affordance pairs uniformly across the full distance spectrum and categorize them into three groups. For training images used as pseudo novel concepts, we generate 10 images per pair using DALL-E~\cite{ramesh2021zeroshottexttoimagegeneration} with GPT-4o~\citep{openai2024gpt4ocard} generated captions that describe different novel designs integrating the specified affordances (Details are described in the Appendix~\ref{app:data_gen}). We then filter images using CLIP similarity scores and manually select the top three. This curriculum-based training enables T2I models to learn basic concept-affordance associations while fusing affordances into coherent and functionally meaningful designs. Thus, the T2I models can successfully produce novel concepts that are visually distinctive and functionally coherent. 


\subsection{Affordance-based Curriculum Learning}
\label{sec:ft}
The goal of fine-tuning T2I models is to enable them to fuse multiple affordances into a single, functionally coherent concept while ensuring visual novelty. With our curriculum, we propose a curriculum learning strategy to fine-tune the diffusion-based T2I models. From a data-driven perspective, training with affordance pairs and DALL-E-generated pseudo-novel concepts helps the model design novel concepts given specified affordances.

To further enhance visual novelty, we incorporate contrastive learning objectives, ensuring that generated images not only reflect desired affordances but also differ from existing concepts associated with them. Specifically, we define two sets of constraints derived from our ontology to guide the model: (1) \textit{Positive Constraints} specify the target affordances that must be included in the novel concepts, shaping their functional structure; (2) \textit{Negative Constraints} consist of all existing concepts from our ontology that already have the target affordances in the positive constraints. These act as references to avoid. By adhering to these constraints, the model generates concepts that successfully integrate the specified affordances while maintaining a high degree of novelty. 
\paragraph{Training Objectives.}
The training objective of fine-tuning is formulated using a triplet loss, which can balance two components to achieve the desired outcomes. The first component aims to minimize the similarity loss between the generated image and the pseudo-novel image created during curriculum construction, ensuring visual novelty. To reduce the overfitting problem, we also sample multiple pseudo-novel images that describe different concepts. Given the set of affordances $\mathcal{A}_{pos} = \{a_1, \cdots, a_n\}$ in the positive constraints, together with a sampled image $I_i^{+}$ from the pseudo-novel images from DALL-E, the positive loss is defined as follows:
\begin{equation}
    \mathcal{L}_{pos}(\theta_{t}) = \| I_{i}^{+} - \hat{I}_{i} \|^2_2 + \mathbb{E}_{ \epsilon, t} \left[ \|\epsilon - \epsilon_\theta(t)\|^2_2 \right],
\end{equation}
where $\theta_{t}$ is T2I model parameters, $\hat{I}_{i}$ denotes the generated image, $\epsilon$ is Gaussian random noise. We employ noise prediction loss, where the model takes the latent embedding of $I_{i}^{+}$ as input and predicts the noise as $\epsilon_\theta(t)$, preventing catastrophic forgetting of learned training distribution.

The second component of the triplet loss maximizes the similarity loss between the generated image and a randomly sampled existing concept image $I_{i}^{-}$ that contains partial affordances from the positive constraints as follows:
\begin{equation}
    \mathcal{L}_{neg}(\theta_{t}) = \| I_{i}^{-} - \hat{I}_{i} \|^2_2
\end{equation}
In this way, the model learns to avoid generating existing concept images and increase its novelty.

Our overall triplet loss is defined as follows:
\begin{equation}
    \mathcal{L}(\theta_{t}) = \mathcal{L}_{pos}(\theta_{t}) - \gamma * \mathcal{L}_{neg}(\theta_{t}),
\end{equation}
where $\gamma$ is an adjustable hyperparameter.
By balancing two losses, our framework ensures that the generated images align with the desired affordances while remaining distinct from existing concepts.

\subsection{Novel Concept Generation during Inference}
\label{sec:inference}
After fine-tuning the diffusion-based T2I models, our approach requires only the desired affordances as positive constraints during inference time, eliminating the need for manually collecting existing concepts as negative constraints. This efficiency gain stems from incorporating both positive and negative constraints--derived from our hierarchical concept ontology--into the training objective. By embedding these constraints during training, the model learns concept-affordance associations and improves its ability to compose parts associated with desired affordances into a novel design. Therefore, the model can produce novel, structured designs without redundant textual prompting.

\section{Experiments}
\label{sec:experiment}



\subsection{Experimental Setup}

\paragraph{Datasets.}
To train T2I models with our approach, we construct a dataset from two types of resources (more details in the Appendix \ref{app:dataset}): (1) \textit{Existing Concept Images}: For each existing concept in our ontology, we collect 60 images from external platforms including Google Images and iStock. To ensure that images are object-centric and aligned with the concept, we filter out low quality images using the CLIP model~\citep{radford2021learningtransferablevisualmodels}. (2) \textit{Generated Novel Concept Images}: With our affordance sampling, we uniformly sample 600 affordance pairs among 235K possible pairs for fine-tuning. For the test dataset, we select 500 affordance pairs among the ones not used for fine-tuning.

\paragraph{Baselines Methods.}
We compare our proposed method against three baseline methods, which are Stable Diffusion~\cite{esser2024scalingrectifiedflowtransformers}, Kandinsky~\cite{arkhipkin2023kandinsky}, and ConceptLab~\cite{richardson2024conceptlab}. While Stable Diffusion and Kandinsky are general T2I models, ConceptLab optimizes generation over diffusion before creative concept design. For a fair comparison, we fine-tune ConceptLab using the same training data as our method. In contrast, our framework directly fine-tuned the diffusion model, integrating the hierarchical visual ontology to enforce the design of a single, coherent concept that achieves multiple affordances. Details on the baselines can be found in the Appendix~\ref{app:baseline}.
\begin{table*}[t]
    \centering
    \resizebox{\textwidth}{!}{%
        \begin{tabular}{ccccccccc}
        \midrule
        & \multicolumn{4}{c}{Automatic Evaluation} & \multicolumn{4}{c}{Human Evaluation} \\
        \cmidrule(lr){2-5} \cmidrule(lr){6-9}
        \textbf{Model} & \textbf{Faithfulness}  & \textbf{Novelty} & \textbf{Practicality} & \textbf{Coherence} & \textbf{Faithfulness}  & \textbf{Novelty} & \textbf{Practicality} & \textbf{Coherence} \\
        \midrule
        Stable Diffusion & $3.77$ & $3.74$ & $3.34$ & $3.29$ & $2.96$ & $2.44$ & $3.02$ & $2.75$  \\
        Kandinsky3 & $3.38$ & $4.02$ & $2.92$ & $3.89$ & $2.95$ & $2.98$ & $3.01$ & $3.41$ \\
        ConceptLab& $3.39$ & $4.08$ & $2.93$ & $3.96$ & $2.73$ & $3.11$ & $2.68$ & $3.54$ \\
        \midrule
        \textsc{Synthia} (Ours) & $\textbf{3.99}$ & $\textbf{4.55}$ & $\textbf{3.35}$ & $\textbf{4.81}$ & $\textbf{3.81}$ & $\textbf{3.89}$ & $\textbf{3.38}$& $\textbf{4.06}$ \\
        \bottomrule
    \end{tabular}}
    \caption{\textbf{Results of the automatic evaluation and human evaluation.} We compare our method with baseline models. Each metric ranges from 0 to 5, where a higher score indicates a better performance. }
    \vspace{-0.1in}
    \label{tbl:abs_results}
\end{table*}

\paragraph{Implementation Details.}
We adopt Kandinsky3.0 \citep{vladimir-etal-2024-kandinsky} as the T2I backbone model; it generates images based on a given text prompt, with an optional negative text prompt to refine outputs. During fine-tuning, we incorporate the desired affordances as positive inputs, while using the existing concepts from the ontology as negative constraints. During inference, we provide only text prompts with desired affordances, ``\textit{a new design that has functions of \{desired affordances\}.}''. Training details are provided in the Appendix \ref{app:hyperparam}


\subsection{Evaluation Metrics}
\label{sec:eval}
\paragraph{Automatic Evaluation.}
We adopt the LLM-as-a-Judge~\citep{zheng2023judging, kim2023prometheus} for automatic evaluation, due to its strong alignment with human judgments. We especially utilize a proprietary model, GPT-4o~\citep{openai2024gpt4ocard}, capable of processing both visual and textual inputs. Building on this, we design four novel metrics to automatically assess the quality of the generated data: \looseness=-1

\begin{itemize}
    \item \textbf{Faithfulness:} This metric evaluates how well the generated object aligns with instructions, focusing on its intended affordances and whether the image effectively conveys the object's purpose.
    \vspace{-0.01in}
    \item \textbf{Novelty} assesses the originality and creativity of the generated design, emphasizing uniqueness and unconventional concepts that surprise or intrigue users.
    \vspace{-0.01in}
    \item \textbf{Practicality} evaluates the real-world applicability of the design. It examines usability, alignment with human preferences, and feasibility for production.
    \vspace{-0.01in}
    \item \textbf{Coherence} evaluates whether the generated image is object-centric, depicting a single clear and functional object without unintended elements. It examines whether multiple affordances are fused into a unified concept rather than shown as separate objects.
\end{itemize}

For all four metrics, we use absolute scores ranging from 1 to 5, with higher scores indicating better quality. However, since the scores for these metrics may be influenced by subjective interpretation, we also include a relative evaluation. Specifically, we present each generated image with its corresponding DALL-E generated image, and ask the automatic evaluator to compare and determine which is superior or if they are equally strong. This relative comparison ensures a more fair evaluation and reduces potential biases (detailed in the Appendix~\ref{appendix:eval_prompt}).
\paragraph{Human Evaluation.}
To assess the quality of the generated concepts beyond automated evaluations, we conduct a human evaluation with 36 non-design expert annotators. Recruited from the university across diverse majors, they are provided with a detailed rubric using the same metrics and a 1-5 scale as automated evaluations. We randomly sample 10 affordance pairs for four models, with each sample independently evaluated. This allows direct comparisons between human and automated scores, capturing nuanced aspects of evaluation quality. Details of the evaluation process are documented in the Appendix~\ref{app:humanevaluation} to ensure transparency.

\begin{figure*}
    \centering
    \includegraphics[width=0.9\textwidth]{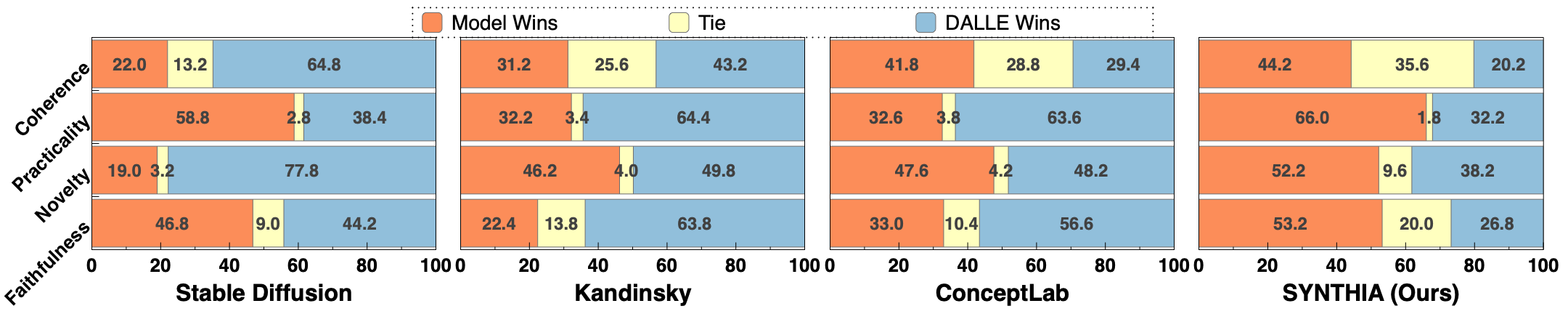}
    \vspace{-0.05in}
    \caption{\textbf{Results of the relative automatic evaluation.} We compare the quality of concepts generated from our models and baselines with ones generated from our data generation pipeline (\S\ref{sec:data_gen}). Numbers indicate the percentage (\%) of baseline model wins, ties, and DALL-E model wins.}
    \label{fig:rel_eval_gt}
    \vspace{-0.07in}
\end{figure*}

\subsection{Results and Analysis}
\paragraph{Automatic Evaluation.}
\label{sec:automatic}
We compare \synthia against three existing T2I models using our evaluation metrics. To ensure a fair comparison, we randomly sample 500 test affordance pairs not seen during training. As shown in the absolute evaluation results (Table \ref{tbl:abs_results}), Stable Diffusion consistently achieves high practicality but lower novelty. This aligns with our observation that it tends to generate existing concepts rather than novel ones (Fig.~\ref{fig:examples}). When it fails to retrieve an existing concept that satisfies all the affordances in the prompt, it often generates multiple disjoint objects without meaningful fusion, leading to lower coherence scores for both concept and affordance levels. In contrast, Kandinsky-3 and ConceptLab exhibit improved novelty and coherence compared to Stable Diffusion, but at the cost of reduced practicality. Their outputs are often more imaginative but less grounded in affordances. Our method, \synthia, outperforms all baselines in faithfulness, novelty, and coherence, while also maintaining strong performance in practicality. These findings are reinforced by the relative evaluation results (Fig.~\ref{fig:rel_eval_gt} and \ref{fig:rel_eval_ours}), where \synthia consistently achieves higher win rates against DALL-E across all metrics compared to baselines. Notably, \synthia even surpasses DALL-E in overall preference, indicating that its generated concepts are perceived as more innovative and functionally coherent. These results demonstrates that fine-tuning with our curriculum strategy enables \synthia to better follow textual instructions, effectively integrate multiple affordances, and generate high-quality novel concepts.
\paragraph{Human Evaluation.}
\label{sec:human_eval}
To assess the consistency of human evaluations, we compute inter-annotator agreement (IAA) between two independent raters, considering ratings to be in agreement if their absolute difference is $\leq 1$.  Overall IAA across all images was 67.5\%, with a Cohen’s Kappa of 22.3\%.  Among the evaluation criteria, Novelty shows the highest agreement (70.9\%), followed by Faithfulness (68.5\%), Practicality (66.4\%), and Coherence (64.1\%) (Fig. \ref{fig:human_eval_results}). Additionally, the agreement between aggregated human ratings and automatic evaluations reached 91.25\%, suggesting strong alignment between human judgments and our automated evaluation framework. These results reinforce that human annotators demonstrate a reasonable level of consistency, and that our automatic evaluation metrics reliably reflect human preferences. Human evaluation further confirms that our model consistently generates functionally coherent and visually novel concepts, achieving superior performance across all metrics (Table~\ref{tbl:abs_results}).

\subsection{Ablation Studies}
\paragraph{The Size of Training Data.}
\begin{figure}[t]
    \centering
    \includegraphics[width=\linewidth]{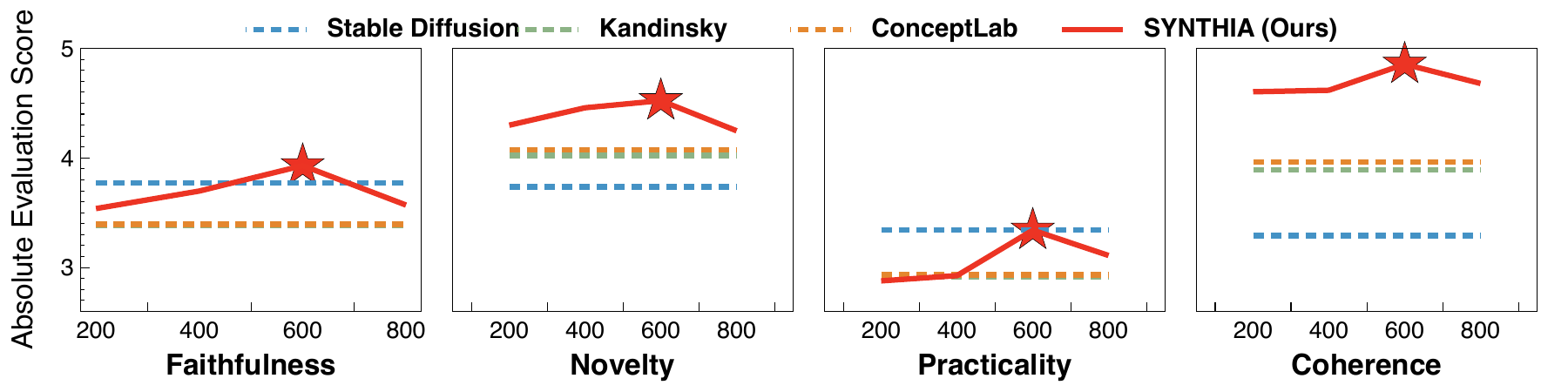}
    \vspace{-0.2in}
    \caption{\textbf{Ablation with different number of training data.} We show the absolute automatic evaluation results of \synthia trained with different number of data.}
    \vspace{-0.1in}
    \label{fig:num_of_data_main}
\end{figure}

In our experiment, we fine-tune the diffusion model using 600 affordance pairs as training data. To investigate the impact of training data size, we compare model performance across different dataset scales: 200, 400, 600, and 800 affordance pairs, evaluated using our automatic evaluation protocol. As shown in Fig. \ref{fig:num_of_data_main}, performance improves as the dataset size increases, reaching its peak at 600 pairs. Notably, our model consistently outperforms all baseline methods across all training sizes, demonstrating the efficacy of our approach.
\paragraph{Effectiveness of Affordance Sampling.}
To examine how the distance between affordance pairs affects concept novelty, we sample $100$ test pairs with the closest and $100$ with the farthest distances. The automatic novelty scores for each group, shown in Table \ref{results_automatic_novelty_sampling}, demonstrate that all three baseline methods achieve relatively low novelty scores for closely related affordance pairs, indicating a tendency to replicate existing concepts rather than generating novel designs. In contrast, our method consistently exhibits high novelty across various distances, outperforming all baselines and demonstrating its ability to generate creative and diverse concepts regardless of affordance similarity. \looseness=-1
\paragraph{Effectiveness of Curriculum Learning.}
\begin{figure}[t]
    \centering
    \includegraphics[width=\linewidth]{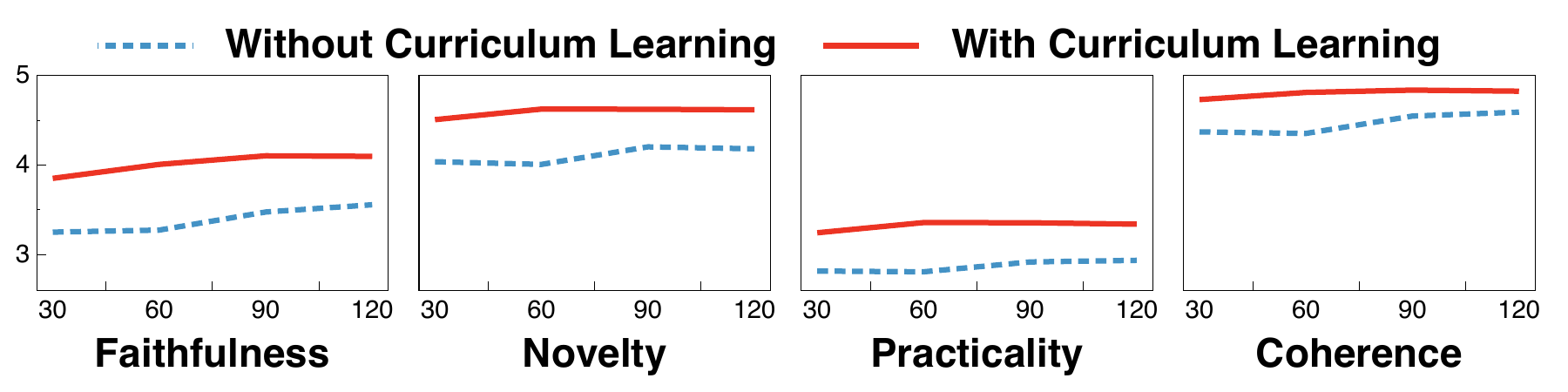}
    \caption{\textbf{Effectiveness of curriculum learning.} We show learning curves of \synthia with different training methods. The X-axis represents training steps.}
    \vspace{-0.08in}
    \label{fig:learning_curve}
\end{figure}
Our framework incorporates a curriculum learning (CL) strategy that gradually increases training difficulty during the fine-tuning process. To assess its effectiveness, we compare model performance with and without CL. Specifically, we train a base T2I model by randomly shuffling the training data and evaluate both settings using the absolute automatic evaluation protocol. As shown in the learning curve (Fig.~\ref{fig:learning_curve}), our CL approach leads to significantly higher scores even in the early stages of training compared to random training, highlighting its role in accelerating learning and guiding the model toward generating high-quality novel concepts. In contrast, training with CL results in a noticeable performance drop (Table \ref{results_automatic_no_curriculum}), further demonstrating the importance of CL in our framework.

\paragraph{Number of Positive Affordances.}
To evaluate the impact of increasing the number of positive affordances in the input prompt, we conduct additional experiments using prompts with three and four affordances. As shown in Tables \ref{results_automatic_triple} and \ref{results_automatic_four}, all methods experience a slight performance drop as input complexity increases. However, \synthia consistently maintains high scores in novelty and coherence, showing strong generalization to more complex affordance combinations---even though it is trained only on two affordance pairs.
This result highlights the scalability of our method in generating high-quality, functionally coherent, and visually novel concepts without additional fine-tuning, showcasing its robustness and generalization capabilities. Moreover, \synthia shows strong performance using only 600 affordance pairs with curriculum learning, accelerating training and enhancing generalization to novel affordance configurations.

\begin{table}[t]
    \centering
    \resizebox{\linewidth}{!}{
        \begin{tabular}{ccccc}
        \toprule
        \textbf{Model} & \textbf{Faithfulness}  & \textbf{Novelty} & \textbf{Practicality} & \textbf{Coherence}\\
        \midrule
        Stable Diffusion & $3.71$ & $3.78$ & $3.37$ & $3.37$ \\
        Kandinsky3 & $3.28$ & $4.07$ & $3.08$ & $4.08$ \\
        ConceptLab & $3.33$ & $4.57$ & $3.03$ & $4.04$\\
        DALL-E & $ \textbf{4.08}$ & $4.32$ & $3.32$ & $4.56$ \\
        \midrule
        \textsc{Synthia} (Ours) & $4.01$ & $\textbf{4.57}$ & $\textbf{3.41}$ & $\textbf{4.78}$\\
        \bottomrule
    \end{tabular}
    }
    \caption{\textbf{Evaluation Results with GPT-4o-mini model.}}
    \vspace{-0.2in}
    \label{tbl:abl_models_main}
\end{table}

\paragraph{Evaluation Model Selection.}
We further conducted experiments using different models, especially GPT-4o-mini, applying the same evaluation prompts and detailed criteria (Table \ref{tbl:abl_models_main}). We observe consistent agreement in the results across different GPT-4o checkpoints, with our model consistently outperforming existing baselines across all metrics, even surpassing DALL-E model in novelty, practicality, and coherence, where DALL-E achieves 4.27, 3.21, and 4.52, respectively, when evaluated with GPT-4o-2024-08-06 (Faithfulness 4.02). \looseness=-1

%

\section{Conclusion}
In this work, we take a step toward affordance-aware generative design by tackling a critical gap in existing text-to-image models---their inability to synthesize visually novel and functionally meaningful designs. With our model, \synthia, we move beyond stylistic variation to purposeful creation, incorporating a hierarchical concept ontology with curriculum-guided fine-tuning to teach models how to compose affordances into unified, novel concepts. Our proposed four-dimensional evaluation metrics capture how well generated concepts align with real-world design principles. Through both human and automatic evaluations, \synthia demonstrates clear superiority in designing imaginative but functionally grounded concepts. By bridging creativity with utility, this work advances AI-driven generative design.

\section*{Limitations}
Our work tackles an important yet underexplored problem of retaining functional coherence in AI for design using T2I models. While our model, in comparison to other state-of-the-art models, is able to generate more coherent and faithful images provided a set of affordances, e.g., \texttt{brew}, \texttt{cut} as in Fig.~\ref{fig:close}, our work inherently relies on the human intuition to evaluate the novelty of the generated concepts. Although we try to alleviate the human bias and lack of coverage using LLM-as-a-judge for automatic evaluation, the question may persist. Moreover, although our concept ontology covers many different concept categories, it does not cover every plausible concept category in the real-world. It would be interesting to see follow-up works explore the direction of constructing a more diverse, richer concept ontology, which in turn would contribute to the generation of more novel concept designs. \looseness=-1


\section*{Acknowledgement}
This research is based upon work supported by U.S. DARPA ECOLE Program No. \#HR00112390060 and an Office of Naval Research (ONR) award \#N00014-23-1-2780. The views and conclusions contained herein are those of the authors and should not be interpreted as necessarily representing the official policies, either expressed or implied, of DARPA, or the U.S. Government. The U.S. Government is authorized to reproduce and distribute reprints for governmental purposes notwithstanding any copyright annotation therein. \looseness=-1

\clearpage
\bibliography{reference}
\clearpage
\appendix
\onecolumn

\begin{center}{\bf {\LARGE Supplementary Materials}}\end{center}
\vspace{0.4in}

\section{Hierarchical Concept Ontology Structure}
\begin{figure}[ht]
    \centering
    \includegraphics[width=0.8\linewidth]{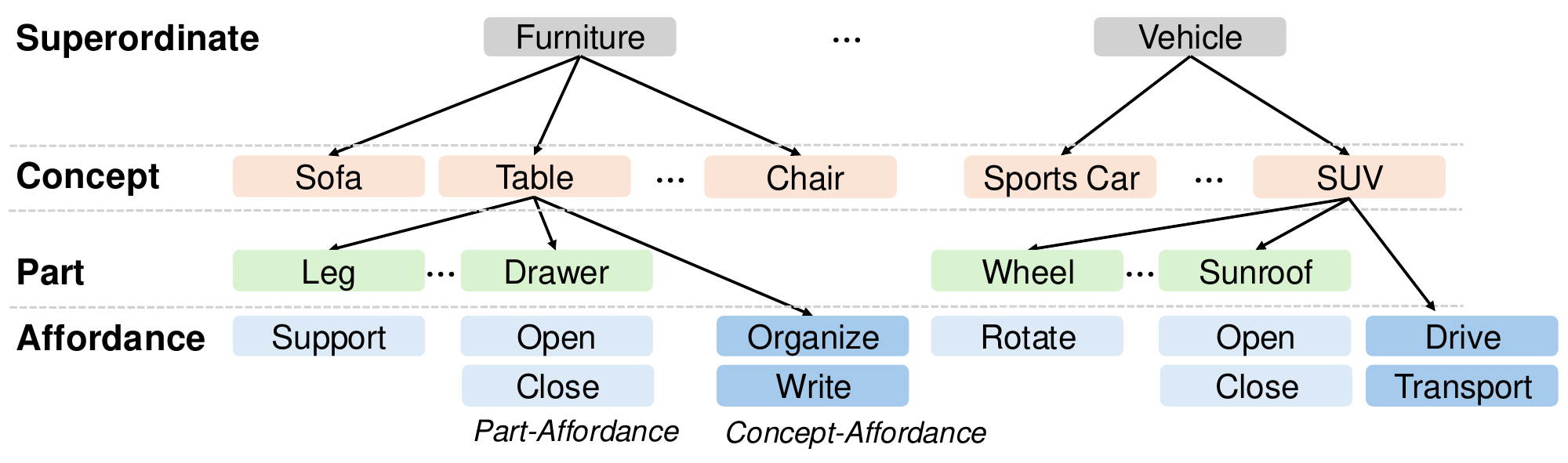}
    \caption{{Hierarchical Concept Ontology.}}
    \label{fig:ontology}
\end{figure}

\section{Experiment Details}
\subsection{Data Generation}
\label{app:data_gen}
To construct high-quality novel concept designs for each sampled affordance pair, we first generate candidate descriptions that satisfy the desired functionalities while ensuring visual novelty relative to existing concepts. Specifically, we prompt GPT-4o~\citep{openai2024gpt4ocard} to produce 10 diverse image captions for each affordance pair, describing novel concepts that fulfill the given affordances and differ in appearance from known objects. We then use these captions to generate 10 corresponding images using DALL-E~\citep{ramesh2021zeroshottexttoimagegeneration}, resulting in a set of pseudo novel concepts.
\vspace{0.1in}
\begin{tcolorbox}[colback=gray!10, colframe=black, title=Prompt for Image Caption Generation]
You are a creative assistant who designs diverse novel concepts satisfying given conditions and generates a description of the concept. 
You should design three different novel concepts where each has all functions in the given positive constraints while the concept is different from the given negative constraints. 

Generate three different descriptions of three novel concepts that contain visible unique characteristics to use generated descriptions as image captions to generate images. Each description should consist of at most three sentences and contain given positive constraints but should not contain non-visible characteristics such as sound, smell, and taste. You must not simply combine multiple existing concepts that have each function but creatively design a single concept that has multiple functions at once. Generate three descriptions of three novel concepts that are not similar to each other but distinct, and each description should be clear without unnecessary explanations for generating images. Please separate each description with \textbackslash n\textbackslash n. Simply follow the format given in the example below.
\begin{verbatim}
{
    Positive Constraints: [sit, store]
    Negative Constraints: [chair, car, sofa, bench, shelve, drawer]
    Image Captions: [“...”]
}
\end{verbatim}
\end{tcolorbox}

\clearpage
\subsection{Dataset}
\label{app:dataset}
To evaluate the performance of our proposed method, we conduct our experiments by constructing a dataset from two types of resources:
\begin{itemize}
    \item \textbf{Existing Concept Images}: For each existing concept in our ontology, we collect a dataset of 60 images from external platforms including Google Images and iStock. To ensure that the dataset is object-centric and minimizes noise, we filter out low-quality images using CLIP model~\citep{radford2021learningtransferablevisualmodels}. Specifically, we compute the similarity between the image embeddings and the text embeddings of the "\texttt{a photo of \{concept name\}}", selecting top-5 images with the highest similarity scores for each concept used as negative constraints. 
    \item \textbf{Generated Novel Concept Images}: With our affordance sampling, we uniformly sample 600 affordance pairs among 82K possible pairs for fine-tuning. For the test dataset, we select 500 affordance pairs among the ones not used for fine-tuning. We use the generated images from the sampled affordances for fine-tuning and evaluation.
    The overall statistics can be found in Table ~\ref{app:dataset_stats}.
\end{itemize}
\begin{table}[ht]
    \centering
    \begin{tabular}{cccc}
    \midrule
    \multirow{2}{*}{\textbf{Dataset}} & \multicolumn{2}{c}{Fine-tuning} & Inference \\
    \cmidrule{2-3}
    & Existing & Generated & Generated \\
    \midrule
    \textbf{\# Concepts} & 772 & 600 & 500 \\
    \textbf{\# Images} & 3860 & 1800 & 500 \\
    \bottomrule
    \end{tabular}%
    \caption{Statistics of the datasets.
    }
    \label{app:dataset_stats}
\end{table}

\subsection{Baseline Methods}
\label{app:baseline}
\begin{itemize}
    \item \textbf{Stable Diffusion} \citep{esser2024scalingrectifiedflowtransformers} is a strong baseline model for high-fidelity image synthesis, which is built on a diffusion-based framework. In this work, we leverage the pre-trained \textit{stable-diffusion-3.5-large} model as the foundation model for the text-to-image task to generate a novel concept. Due to the limited context window length, we input only the positive constraints that contain the desired affordances, omitting the negative constraints associated with existing concepts.
    \item \textbf{Kandinsky} \citep{arkhipkin2023kandinsky, vladimir-etal-2024-kandinsky} model serves as another strong baseline model for comparison. We utilize the pre-trained \textit{Kandinsky 3.0} model without any finetuning as the baseline, aligning it with the foundation model used in our proposed framework. This approach ensures a consistent starting point for evaluation and a fair comparison. This baseline allows us to effectively demonstrate the impact of our proposed training framework by comparing performance before and after the fine-tuning process.
    \item \textbf{ConceptLab} \citep{richardson2024conceptlab} is a state-of-the-art framework designed for creative concept generation, which leverages an innovative approach that formulates the generation problem as an optimization process over the output space of the diffusion prior. It adopts a similar input format to the one used in our setting. We follow the finetuning process in the original framework, applying our training data to generate novel concepts. We then compare the generation quality during the inference time.
\end{itemize}

\subsection{Hyperparameter Settings}
\label{app:hyperparam}
To fine-tune the diffusion model with the constructed dataset from DALL-E,
we use the optimizer AdamW \citep{loshchilov2019decoupledweightdecayregularization} with a learning rate of $\eta = \{5 * 10^{-6}, 10^{-6}, 5 * 10^{-7}\}$ to avoid the catastrophic forgetting problem. During the curriculum learning, we split the dataset into three groups based on their difficulty and trained 20 epochs for each group. For the weight factor in the triplet loss, we set $\gamma = \{0, 0.2, 0.5, 0.8, 1\}$. We finetune the UNet part in the pre-trained model and freeze the weights of other components in the diffusion model. The training takes about 1 GPU hour with NVIDIA A100 GPU.

\subsection{Automatic Evaluation}
\label{appendix:eval_prompt}
\subsubsection{Absolute Automatic Evaluation}
\vspace{0.1in}
\begin{tcolorbox}[colback=gray!10, colframe=black, title=Prompt for Absolute Automatic Evaluation, breakable]
Please act as an impartial evaluator to assess the quality of a single concept image generated by an AI, based on the user’s requirements. Your evaluation should use the following three criteria, each scored on a scale of 1 to 5:\\

Faithfulness: Evaluate how well the object aligns with the provided instructions, including its intended affordances and functionalities. Does the text and image together indicate that the object serves the purpose for which it was designed?\\
Scoring:\\
5: Flawlessly combines all specified functionalities as per the instructions. Text and image work in harmony to demonstrate a well-designed and fully functional object.\\
4: Fulfills most instructions and intended functionalities, with only minor inconsistencies or missing details. The text and image are mostly aligned.\\
3: Partially fulfills the instructions. Some functionalities are present but not well-integrated or consistent. There may be a minor mismatch between text and image.\\
2: Struggles to meet the provided instructions, missing key functionalities or combining them poorly. Text and image may conflict.\\
1: Does not follow the instructions at all. Functionalities are completely missing, irrelevant, or nonsensical.\\

Novelty: Assess the originality and innovation of the design. Does the object show an exciting, novel design that would surprise or intrigue users?\\
Scoring:\\
5: Highly innovative, unique, and impressive. Inspires curiosity or excitement, making it highly desirable to explore.\\
4: Contains interesting and novel elements, showing clear creative thought and appeal.\\
3: Displays moderate novelty, with some unique features but remaining relatively conventional or uninspiring.\\
2: Shows limited novelty, with minimal creativity and overly simplistic or derivative design.\\
1: Entirely unoriginal and mundane, lacking any creativity and appearing common or uninspiring.\\

Practicality: Evaluate the real-world applicability of the object. Does the design make sense for human use? Would it align with human preferences and be feasible for production?\\
Scoring:\\
5: Extremely practical and human-centric. Highly functional, aligns perfectly with human preferences, and seamlessly fits into real-world scenarios.\\
4: Mostly practical and applicable, with minor limitations that could be addressed to improve usability.\\
3: Somewhat practical but with notable flaws or unrealistic elements that may limit usability in real-world scenarios.\\
2: Largely impractical, with significant flaws or inconsistencies that make it unlikely to be useful.\\
1: Entirely impractical and unusable, failing to align with human preferences or real-world feasibility.\\

Coherence: This metric evaluates whether the image generated by the AI model contains only one primary object as instructed, focusing on the object's clarity and functionality without the presence of additional, unintended objects.\\
Scoring:\\
5: The image perfectly showcases one distinct object that aligns with the described attributes. There are no extraneous objects or elements that distract from the main object.\\
4: The primary object is clear and mostly isolated, but there may be minor elements in the background or periphery that do not significantly detract from the main object.\\
3: The main object is present and identifiable, but there are other elements in the image that somewhat distract from its clarity and functionality.\\
2: The image contains multiple objects where the main object is not clearly dominant or distinguishable from other unnecessary elements.\\
1: The image primarily features multiple objects, making it difficult to identify the intended single object; the composition is cluttered or entirely irrelevant to the instruction.

Provide a score for each criterion, followed by a concise explanation justifying your ratings. Your final response should strictly follow this format: 
\begin{verbatim}
{
    "Faithfulness": [Your Faithfulness Score],
    "Novelty": [Your Novelty Score],
    "Practicality": [Your Practicality Score],
    "Coherence": [Your Coherence Score]
}
\end{verbatim}
\end{tcolorbox}

\vspace{0.1in}
\subsubsection{Relative Automatic Evaluation}
\label{app:rel_eval_detail}
\vspace{0.1in}
\begin{tcolorbox}[colback=gray!10, colframe=black, title=Prompt for Relative Automatic Evaluation]
Please act as an impartial evaluator to assess the quality of concept images generated by two AI concept generators based on the user’s requirements. The evaluation criteria are as follows:
Faithfulness: Evaluate how well the object aligns with the provided instructions, including its intended affordances and functionalities. Does the text and image together indicate that the object serves the purpose for which it was designed? 
Novelty: Assess the originality and innovation of the design. Does the concept demonstrate a surprising or intriguing approach that stands out as fresh and exciting?
Practicality: Evaluate the real-world applicability of the concept. Does the design make sense for human use, align with user preferences, and appear feasible for production?
Coherence: This metric evaluates whether the image generated by the AI model contains only one primary object as instructed, focusing on the object's clarity and functionality without the presence of additional, unintended objects.
Provide your answer based on the following available choices:
"A" if the first image is better,
"B" if the second image is better,
"C" if both are equally strong.
Your final response should strictly follow this format: 
\begin{verbatim}
{
    "Faithfulness": [Your Faithfulness Choice],
    "Novelty": [Your Novelty Choice],
    "Practicality": [Your Practicality Choice],
    "Coherence": [Your Coherence Choice]
}
\end{verbatim}
\end{tcolorbox}

\clearpage
\subsection{Human Evaluation}
\label{app:humanevaluation}
\begin{tcolorbox}[colback=gray!10, colframe=black, title=Human Evaluation Instruction, breakable]
\textbf{Objective}

The goal of this evaluation is to assess the quality of novel concepts that integrate multiple affordances into a single, coherent design. Affordance refers to the functional properties of an object or its components. For example, a sofa affords the function of sitting, while its legs provide the function of support.

As an annotator, you will evaluate the given concepts based on four key metrics: \textbf{faithfulness}, \textbf{novelty}, \textbf{practicality}, and \textbf{coherence}. Each metric is defined below, along with its respective scoring criteria. \\

\textbf{Evaluation Criteria}

Faithfulness (Does the concept effectively integrate the specified affordances?)\\
This metric assesses whether the generated concept successfully incorporates all provided affordances in a meaningful and functional manner. \\

Scoring Scale:

5 – Fully integrates all specified affordances, demonstrating a well-designed and fully functional object.

4 – Incorporates all affordances with minor inconsistencies or slight missing details.

3 – Partially fulfills the affordances; some functionalities are present but not well-integrated or consistent.

2 – Struggles to meet the provided affordances; key functionalities are missing or poorly combined.

1 – Does not incorporate the specified affordances; functionalities are entirely missing, irrelevant, or nonsensical. \\

Novelty (To what extent does the concept demonstrate originality and innovation?) \\
This metric evaluates the uniqueness and creative appeal of the design, considering whether it introduces novel elements that would intrigue or surprise users.

Scoring Scale:

5 – Highly innovative and unique; presents a strikingly original concept that is engaging and thought-provoking.

4 – Contains clear novel elements, demonstrating creative thought and originality.

3 – Moderately novel; some unique aspects are present, but the overall concept remains relatively conventional.

2 – Limited novelty; the design appears simplistic, derivative, or lacking in creativity.

1 – Entirely unoriginal and uninspiring, closely resembling existing designs with no innovative aspects. \\

Practicality (Is the design feasible and suitable for real-world use?) \\
This metric assesses whether the concept is functionally viable and aligned with human preferences and usability considerations.

Scoring Scale:

5 – Highly practical and user-centered; seamlessly functional and feasible for real-world applications.

4 – Mostly practical; minor limitations exist but do not significantly hinder usability.

3 – Somewhat practical; contains notable flaws or unrealistic elements that may limit real-world applicability.

2 – Largely impractical; significant design flaws make real-world usability unlikely.

1 – Entirely impractical and non-functional; does not align with human preferences or feasibility constraints.

Coherence (Does the image clearly depict a single, distinct object?) \\
This metric evaluates whether the design presents a singular, well-defined object, free from extraneous elements that may obscure its intended functionality.

Scoring Scale:

5 – The image clearly and exclusively depicts a single object that integrates all specified affordances without any distractions.

4 – The primary object is distinct and well-defined, though minor background elements may be present without significantly detracting from clarity.

3 – The main object is identifiable, but additional elements in the image introduce some visual or conceptual distractions.

2 – The image contains multiple objects, making it difficult to distinguish the intended primary object or missing at least one affordance.

1 – The image primarily features multiple objects, with affordances spread across different elements rather than a unified concept, making it unclear what the primary object is. \\

\textbf{Final Instructions}

Please evaluate each concept independently based on the above criteria.

Assign a score for each metric according to the provided descriptions.

If a concept does not fit neatly into the scoring categories, use your best judgment to determine the most appropriate score.

Your evaluations will contribute to assessing the effectiveness of novel concept generation and help improve future designs. Thank you for your participation.
\end{tcolorbox}

\vspace{0.2in}
\section{Additional Experimental Results}
\subsection{Results of Relative Automatic Evaluation}
\begin{figure}[ht]
    \centering
    \includegraphics[width=\textwidth]{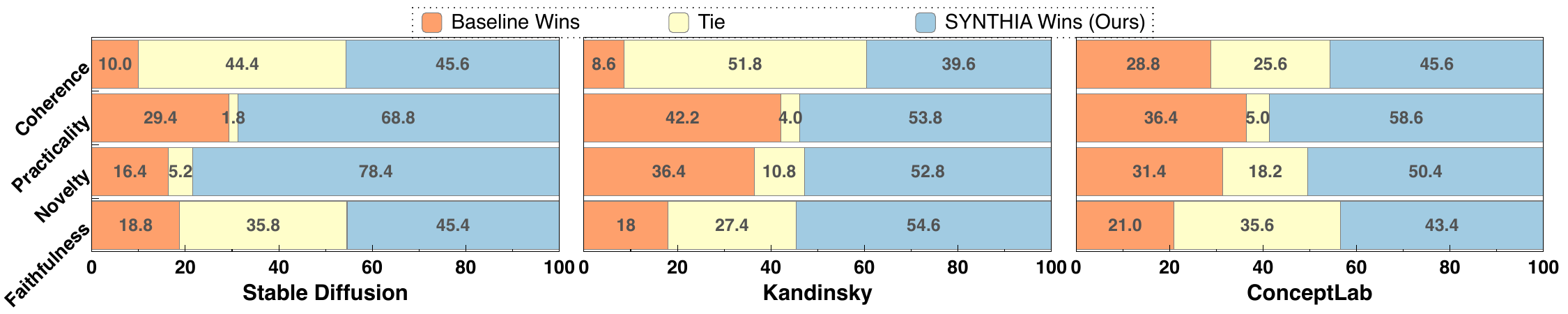}
    \vspace{-0.2in}
    \caption{\textbf{Results of the relative automatic evaluation.} We compare the quality of concepts generated from our models with ones from existing T2I models. Numbers indicate the percentage (\%) of baseline model wins, ties, and \synthia model wins.}
    \label{fig:rel_eval_ours}
\end{figure}

As described in Sec~\ref{sec:automatic}, we perform relative evaluation following the protocol~\ref{app:rel_eval_detail}, comparing each method
s generated concepts against those from DALL-E. We show that \synthia achieves higher win rates than the baselines in all metrics when evaluated against DALL-E. Moreover, we conduct direct pairwise comparisons between \synthia and the baselines. As shown in Fig.~\ref{fig:rel_eval_ours}, \synthia outperforms all baselines across every metric, demonstrating its superior ability to design novel concepts that are both visually novel and functionally coherent, given a set of desired affordances.

\clearpage
\subsection{Results of Human Evaluation}
\begin{figure}[ht]
    \centering
    \includegraphics[width=\textwidth]{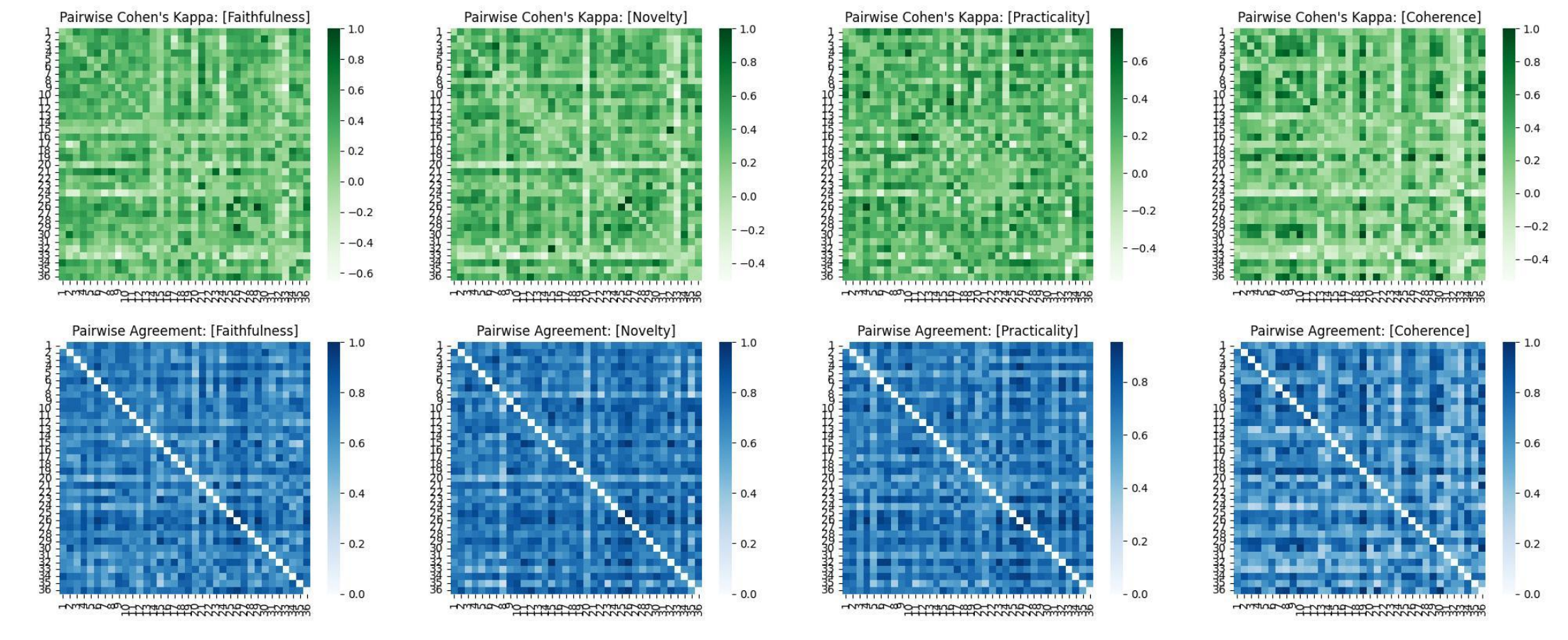}
    \caption{Results of the human evaluation.}
    \label{fig:human_eval_results}
\end{figure}

In addition to the absolute inter-annotator agreement (IAA) reported in Sec~\ref{sec:human_eval}, we further visualize the agreement between individual human annotators. As shown in Fig.~\ref{fig:human_eval_results}, the overall agreement is high, supporting the reliability and validity of our human evaluation results.

\subsection{Effectiveness of Affordance-based Textual Prompt}

During the inference for novel concept generation, we align textual prompts with affordances as we described in Sec~\ref{sec:inference}. To further demonstrate the effectiveness of incorporating affordance to guide the novel concept synthesis, we additionally conduct experiments with textual input based on concept-level prompts without adding additional signals of affordances. We especially use prompts of ``\textit{a new design that includes functions from \{concepts that possess desired affordances -- e.g., sofa, chair\}.}''. For a fair comparison, we use the same fine-tuned models and baseline models but only utilize different prompts (concept-level prompt vs. affordance-level prompt) to evaluate the quality of the generated concepts. As shown in Table~\ref{results_automatic_triple}, affordance-level prompts facilitate a better design across different models, achieving an improvement in faithfulness and coherence. This demonstrates that aligning textual prompts with affordances can benefit even off-the-shelf T2I models to achieve better functionally coherent concepts. However, they are still far behind our models, showing the fundamental challenges in affordance-aware novel concept synthesis. 
\begin{table*}[ht]
    \centering
    \resizebox{0.8\linewidth}{!}{%
    \begin{tabular}{cccccc}
    \midrule
    \textbf{Model} & \textbf{Type} & \textbf{Faithfulness}  & \textbf{Novelty} & \textbf{Practicality} & \textbf{Coherence} \\
    \midrule
    \multirow{2}{*}{Stable Diffusion} & Concept & $2.99$ &  $3.83$ &  $2.87$ & $2.65$ \\
    & Affordance & $3.73$ & $3.80$ & $3.33$ & $3.97$ \\
    \hdashline
    \multirow{2}{*}{Kandinsky3} & Concept & $3.10$ & $3.92$ & $2.79$ & $3.24$ \\
    & Affordance & $3.52$ & $3.91$ & $3.11$ & $4.20$ \\
    \hdashline
    \multirow{2}{*}{ConceptLab} & Concept & $3.61$ & $4.13$ & $3.14$ & $4.31$ \\
    & Affordance & $3.78$ & $4.03$ & $3.08$ & $4.27$ \\
    \midrule
    \multirow{2}{*}{\synthia (Ours)} & Concept & $3.90$ & $4.34$ & $3.28$ & $4.60$ \\
    & Affordance & $3.99$ & $4.45$ & $3.36$ & $4.76$ \\
    \bottomrule
    \end{tabular}%
    }
    \caption{\textbf{Results of the automatic evaluation with three positive affordances.} We compare our method with baselines. For each metric, a higher number indicates a better performance, where the score ranges between 0 and 5.
    }
    \label{results_automatic_triple}
\end{table*}

\begin{table*}[t]
    \centering
    \resizebox{0.8\linewidth}{!}{%
    \begin{tabular}{cccccc}
    \midrule
    \textbf{Model} & \textbf{Type} & \textbf{Faithfulness}  & \textbf{Novelty} & \textbf{Practicality} & \textbf{Coherence} \\
    \midrule
    \multirow{2}{*}{Stable Diffusion} & Concept & $2.74$ &  $3.73$ &  $2.73$ & $2.38$ \\
    & Affordance & $3.41$ & $3.82$ & $3.08$ & $3.71$ \\
    \hdashline
    \multirow{2}{*}{Kandinsky3} & Concept & $2.92$ & $3.88$ & $2.62$ & $2.82$ \\
    & Affordance & $3.33$ & $3.87$ & $3.13$ & $4.07$ \\
    \hdashline
    \multirow{2}{*}{ConceptLab} & Concept & $3.67$ & $4.03$ & $3.03$ & $4.42$ \\
    & Affordance & $3.61$ & $3.98$ & $2.98$ & $4.17$ \\
    \midrule
    \multirow{2}{*}{\synthia (Ours)} & Concept & $3.85$ & $4.36$ & $3.14$ & $4.59$ \\
    & Affordance & $3.86$ & $4.52$ & $3.25$ & $4.80$ \\
    \bottomrule
    \end{tabular}%
    }
    \caption{\textbf{Results of the automatic evaluation with four positive affordances}. We compare our method with baselines. For each metric, a higher number indicates a better performance, where the score ranges between 0 and 5.
    }
    \label{results_automatic_four}
\end{table*}

\vspace{1.0in}
\subsection{Effectiveness of Affordance Sampling}

\begin{table}[ht]
    \centering
    \begin{tabular}{cccc}
    \midrule
    \textbf{Model} & \textbf{Type} & \textbf{Close}  & \textbf{Distant} \\
    \midrule
    \multirow{2}{*}{Stable Diffusion} & Concept & $3.92$ &  $4.13$ \\
    & Affordance & $3.71$ & $3.88$ \\
    \hdashline
    \multirow{2}{*}{Kandinsky3} & Concept & $4.04$ & $4.13$ \\
    & Affordance & $4.18$ & $4.14$ \\
    \hdashline
    \multirow{2}{*}{ConceptLab} & Concept & $4.18$ & $4.17$ \\
    & Affordance & $3.96$ & $4.20$ \\
    \midrule
    \multirow{2}{*}{\synthia (Ours)} & Concept & $4.26$ & $4.33$ \\
    & Affordance & $4.46$ & $4.59$ \\
    \bottomrule
    \end{tabular}
    \caption{\textbf{Results of the automatic novelty evaluation with different distances.} We compare our method with baselines. For each metric, a higher number indicates a better performance, where the score ranges between 0 and 5.
    }
    \label{results_automatic_novelty_sampling}
\end{table}

\vspace{0.1in}
\subsection{Effectiveness of Curriculum Learning}
\begin{table}[ht]
    \centering
    \small
    \resizebox{0.8\linewidth}{!}{%
    \begin{tabular}{cccccc}
    \midrule
    \textbf{Model} & \textbf{Type} & \textbf{Faithfulness}  & \textbf{Novelty} & \textbf{Practicality} & \textbf{Coherence} \\
    \midrule
    \multirow{2}{*}{\textsc{Synthia} without CL} & Concept & $3.61$ & $4.03$ & $3.07$ & $4.30$ \\
    & Affordance & $3.56$ & $4.13$ & $2.95$ & $4.59$ \\
    \midrule
    \multirow{2}{*}{\synthia (Ours)} & Concept & $3.97$ & $4.33$ & $3.30$ & $4.51$ \\
    & Affordance & $3.99$ & $4.55$ & $3.35$ & $4.81$ \\
    \bottomrule
    \end{tabular}%
    }
    \caption{\textbf{Results of the automatic evaluation with and without curriculum learning.} For each metric, a higher number indicates a better performance, where the score ranges between 0 and 5.
    }
    \label{results_automatic_no_curriculum}
\end{table}

\subsection{More Visual Examples}
\begin{figure}[ht]
    \centering
    \includegraphics[width=0.8\linewidth]{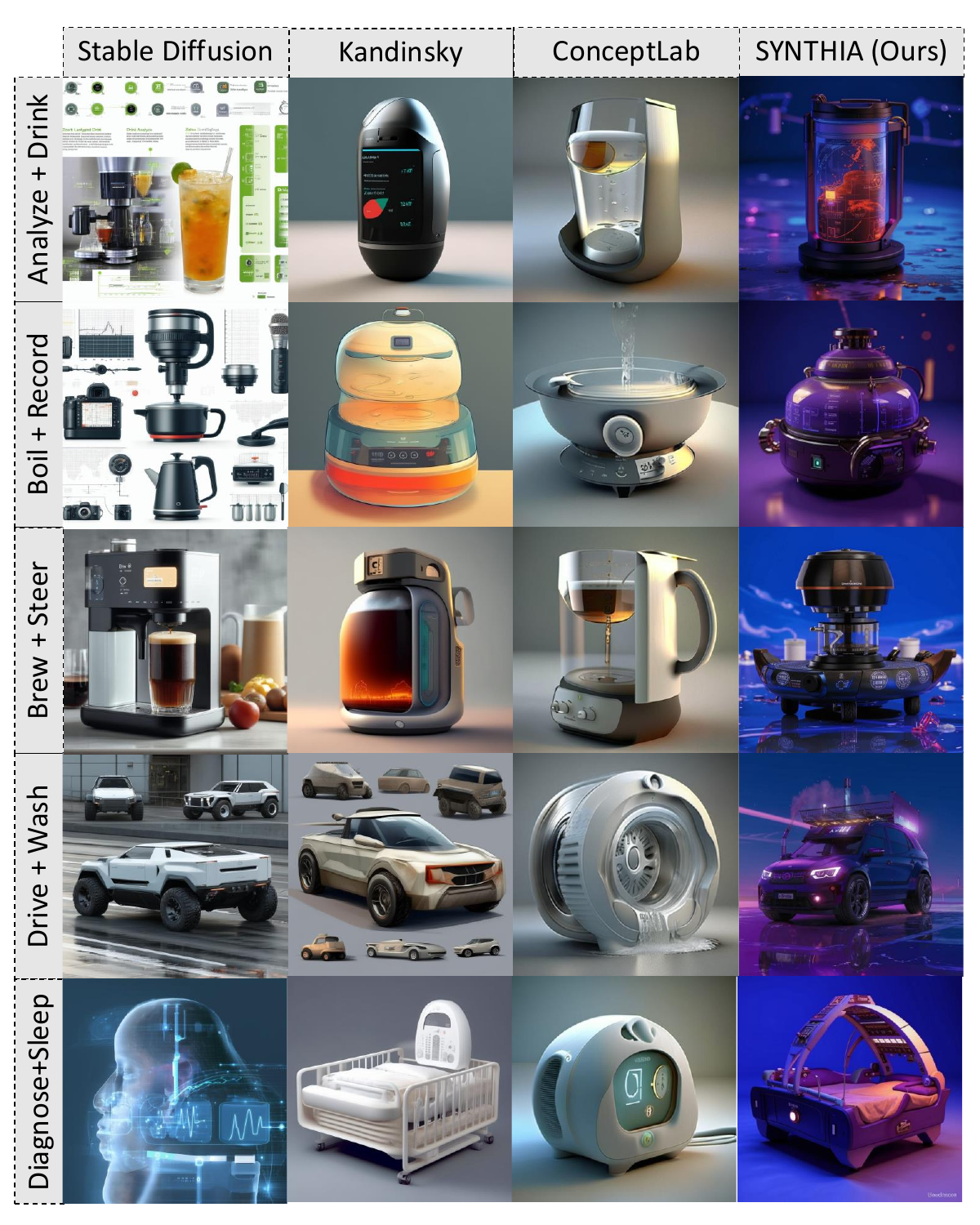}
    \caption{More generated examples from the baseline methods and \synthia (ours). }
    \label{fig:app-visual-eg}
\end{figure}

We include additional visual examples comparing the baseline methods with \textsc{Synthia} (Ours). As illustrated in Figure~\ref{fig:app-visual-eg}, our model outperforms in generating concepts that align closely with the input affordance pairs. Moreover, \textsc{Synthia} consistently produces outputs with higher quality across 4 key dimensions, including faithfulness, novelty, practicality, and coherence.

\section{Relevance to NLP domain}
While our work focuses on Text-to-Image (T2I) generation, it is fundamentally grounded in Natural Language Processing (NLP) domain. Specifically, we leverage NLP methods to extract concept-affordance associations from commonsense knowledge bases and incorporate the structured knowledge into the image generation process. Our interdisciplinary approach not only expands the horizons of text-to-image generation, but also advances core NLP research by demonstrating how linguistic and semantic understanding can inform and enrich multimodal systems. This contribution aligns closely with the NLP community growing interest in multimodal AI and the integration of language with other modalities.

\section*{Ethical Consideration}
We acknowledge that our work is aligned with the \textit{ACL Code of the Ethics} \footnote{\url{https://www.aclweb.org/portal/content/acl-code-ethics}} and will not raise ethical concerns.
We do not use sensitive datasets/models that may cause any potential issues/risks.



\end{document}